\def\year{2021}\relax
\documentclass[letterpaper]{article} 
\usepackage{aaai21}  
\usepackage{times}  
\usepackage{helvet} 
\usepackage{courier}  
\usepackage[hyphens]{url}  
\usepackage{graphicx} 
\usepackage{natbib}  
\usepackage{caption} 

\usepackage{qiangstyle}
\usepackage{subfigure}
\usepackage{multirow}
\usepackage{amsmath}
\usepackage{amsfonts}
\usepackage{xcolor}
\usepackage{makecell}
\newtheorem{theorem}{Theorem}

\newtheorem{definition}{Definition}

\usepackage{tabularx}
\makeatletter
\def\hlinewd#1{%
\noalign{\ifnum0=`}\fi\hrule \@height #1 %
\futurelet\reserved@a\@xhline}
\makeatother

\title{Post-training Quantization with Multiple Points:\\ Mixed Precision without Mixed Precision}
\author {
    
    Xinghcao Liu\textsuperscript{\rm 1},
    Mao Ye\textsuperscript{\rm 1},
    Dengyong Zhou\textsuperscript{\rm 2}, 
    Qiang Liu \textsuperscript{\rm 1} \\
}
\affiliations {
    \textsuperscript{\rm 1} The University of Texas at Austin
    \textsuperscript{\rm 2} Google Brain \\
}
\begin{document}

\global\long\def\Q{\mathcal{Q}}%
\global\long\def\w{\boldsymbol{w}}%
\global\long\def\R{\mathbb{R}}%
\global\long\def\M{\mathcal{M}}%
\global\long\def\bp{\boldsymbol{\phi}}%
\global\long\def\r{\boldsymbol{r}}%
\maketitle

\begin{abstract}
We consider the post-training quantization problem, 
which discretizes the weights of pre-trained deep neural networks without re-training the model. 
We propose \emph{multipoint quantization}, 
a quantization method that approximates a full-precision weight vector 
using a linear combination of multiple vectors of low-bit numbers; 
this is in contrast to typical quantization methods that approximate each weight using a single low precision number. 
Computationally, we construct the multipoint quantization with an efficient greedy selection procedure, and adaptively decides the number of low precision points on each quantized weight vector based on the error of its output.  
This allows us to achieve higher precision levels for important weights that greatly influence the outputs, yielding an ``effect of mixed precision'' but without physical mixed precision implementations  (which requires specialized hardware accelerators~\citep{wang2019haq}). 
Empirically, our method can be implemented by common operands, bringing almost no memory and computation overhead. We show that our method outperforms a range of state-of-the-art methods on ImageNet classification and it can be generalized to more challenging tasks like PASCAL VOC object detection. 
\end{abstract}

\vspace{-10pt}
\section{Introduction}
The past decade has witnessed the great success of deep neural networks (DNNs) in many fields. Nonetheless, DNNs require expensive computational resources and enormous storage space, making it difficult for deployment on  resource-constrained devices, such as devices for Internet of Things (IoT), processors on smart phones, and embeded controllers in mobile robots~\citep{howard2017mobilenets,  xu2018scaling}.

Quantization is a promising method for creating more energy-efficient deep learning systems~\citep{han2015deep, hubara2017quantized, neta_zmora_2018_1297430, cheng2018recent}.
By approximating real-valued weights and activations using low-bit numbers, 
quantized neural networks (QNNs) trained with state-of-the-art algorithms
\citep[e.g.,][]{courbariaux2015binaryconnect, rastegari2016xnor, louizos2018relaxed, li2019additive}   
can be shown to perform similarly as their full-precision counterparts~\citep[e.g.,][]{  
jung2019learning, li2019additive}.   



This work focuses on the problem of \emph{post-training quantization}, which aims to generate a QNN from a 
pretrained full-precision network, without accessing the original training data 
\citep[e.g.,][]{sung2015resiliency, krishnamoorthi2018quantizing, zhao2019improving, meller2019same, banner2019post, nagel2019data, choukroun2019low}. This scenario appears widely in practice. For example, when a client wants to deploy a full-precision model provided by a machine learning service provider in low-precision, the client may have no access to the original training data due to privacy policy. 
In addition, compared with training QNNs from scratch, prost-training quantization is much more efficient computationally.

\emph{Mixed precision} 
is a recent advanced technology to 
boost the performance of QNNs~\citep{wang2019haq, banner2019post, gong2019mixed, Dong_2019_ICCV}. The idea is to assign more bits to important layers (or channels) and less bits to unimportant layers/channels 
to better control the overall quantization error
and balance the
accuracy and cost more efficiently. 
%
The difficulty, however,
is that current mixed precision methods require specialized hardware~\citep[e.g.,][]{wang2019haq}. Most commodity hardware do not support efficient mixed precision computation (e.g. due to chip area constraints~\citep{horowitz20141}). This makes it difficult to implement mixed precision in practice, despite that it is highly desirable.


In this paper, we propose 
\emph{multipoint quantization} for post-training quantization, 
which can achieve the flexibility similar to mixed precision, but uses only a single precision level. 
The idea is to approximate a full-precision weight vector by a linear combination of multiple low-bit vectors. 
This allows us to use a larger number of low-bit vectors to approximate the weights of more important channels, while use less points to approximate the insensitive channels. It enables a flexible trade-off between accuracy and cost at a per-channel basis, while using only a single precision level. Because it does not require physical mixed precision implementation, our method can be easily deployed on commodity hardware by common operands.

We propose a greedy algorithm to iteratively find the optimal low-bit vectors to minimize the approximation error. The algorithm sequentially adds the low-bit vector that yields largest improvement on the error, until a stopping criterion is met. We develop a theoretical analysis, showing that the error decays exponentially with the number of low-bit vectors used. The fast decay of the greedy algorithm ensures small overhead after adding these additional points.

Our multipoint quantization is computationally efficient. The key advantage is that it only involves multiply-accumulate (MAC) operations during inference, which has been highly optimized in normal deep learning devices. We adaptively decide the number of low precision points for each channel by measuring its output error. Empirically, we find that there are only 
a small number of channels that require a large number of points. By applying multipoint quantization on these channels, the performance of the QNN is improved significantly without any training or fine-tuning. Empirically, it only brings a negligible increase of memory cost.

   

We conduct experiments on ImageNet classification  with different neural architectures.
Our method performs favorably against the state-of-the-art methods. 
It even outperforms the method proposed by \citet{banner2019post} in accuracy, which exploits physical mixed precision. We also verify the generalizability of our approach by applying it to PASCAL VOC object detection tasks.

\vspace{-10pt}
\section{Related Works}
Quantized neural networks has made significant progress with training~\citep{courbariaux2015binaryconnect, han2015deep, zhu2016trained, rastegari2016xnor, mishra2017wrpn, neta_zmora_2018_1297430, cheng2018recent, krishnamoorthi2018quantizing, li2019additive}. The research of post-training quantization is conducted for scenarios when training is not available~\citep{krishnamoorthi2018quantizing,meller2019same, banner2019post, zhao2019improving}. Hardware-aware Automated Quantization~\citep{wang2019haq} is a pioneering work to apply mixed precision to improve the accuracy of QNN, which needs fine-tuning the network. It inspired a line of research of training a mixed precision QNN~\citep{gong2019mixed, Dong_2019_ICCV}. 
\citet{banner2019post} first exploits mixed precision to enhance the performance of post-training quantization.

Using multiple binary filters to approximate a full-precision filter has been investigated in supervised learning of binary neural networks~\citep{guo2017network, lin2017towards, zhuang2019structured}. Our work differs from these works in two dimensions: (1) Instead of supervised learning, our work focuses on post-training quantization, where only a small fraction of data can be used. (2) Our work goes beyond binary case. Our method and theory are applicable for quantization with arbitrary bits. This is non-trivial since, with higher bits, the weights are not simply $\pm 1$, leading to a much more difficult optimization problem.
\vspace{-10pt}
\section{Method} 
We start by introducing the background on post-training quantization. Then we discuss the main framework of multipoint quantization, its application to deep neural networks, and its implementation overhead.
\vspace{-10pt}
\subsection{Preliminaries: Post-trainig Quantization}
\label{sec:Preliminaries} 
Given a pretrained full-precision neural network $f$, the goal of post-training quantization is to generate a quantized neural network (QNN) $\tilde{f}$ with high performance. 
We assume the full training dataset of $f$ is unavailable, but there is 
a small
\emph{calibration dataset} $D=\{\vec x_i\}_{i=1}^N$, where $N$ is a very small size, e.g. $N=256$. 
The calibration set is used only for choosing a small number of hyperparameters of our algorithm, and we can not directly train $\tilde f$ on it because it is too small and would cause overfitting.

The $b$-bit linear quantization amounts to approximate real numbers using the following quantization set $\setvv Q$, 
\begin{align}
   \setvv Q = K\times [-1: \epsilon_b: 1] ~+~ B,  &&
   \epsilon_b := \frac{1}{2^{b-1} -1}, 
\end{align}
where $[-1: \epsilon_b: 1]$ denotes the uniform grid on $[-1,1]$ with increment $\epsilon_b$ between elements, and $K > 0$ is a scaling factor that controls the length of $\setvv Q$ and $B$ specifies center of $\setvv Q$.  


Then we map a floating number $t$ to $\setvv Q$ by, 
\begin{equation}
    \tilde t  = 
    [t]_{\setvv Q}:=\argmin_{z\in \setvv Q} |t -z|, 
\end{equation}
where $[\cdot]_{\setvv Q}$ denotes the nearest rounding operator w.r.t.  $\setvv Q$. For a real vector $\vec t = \left(t_1, t_2, \dots, t_d\right) \in \RR^d$, we map it to $\setvv Q^d$ by, 
\begin{equation}
\label{eq:nnquant}
    \tilde{\vec t} = [\vec t]_{\setvv Q} = ([t_1]_{\setvv Q}, [t_2]_{\setvv Q}, \dots, [t_d]_{\setvv Q}).
\end{equation}
Further,
$[\cdot]_{\setvv Q}$ can be generalized to higher dimensional tensors by first stretching them to one-dimensional vectors then applying Eq.~\ref{eq:nnquant}. 

Since all the values are larger than $K$ (or smaller than $-K$) will be clipped, $K$ is also called
the \emph{clipping factor}. Supposing $\setvv Q$ is used to quantize vector $\vec t$, a naive choice of $K$ is the element with the maximal absolute value in $\vec t$. In this case, no element will be clipped.
However, because the weights in a layer/channel of a neural network empirically follows a bell-shaped distribution, properly shrinking $K$ can boost the performance. Different clipping methods have been proposed to optimize $K$~\citep{zhao2019improving}.

There are two common configurations for {post-training quantization}, 
\emph{per-layer quantization} and 
\emph{per-channel quantization}.
Per-layer quantization assigns the same $K$ and $B$ for all the weights in the same layer. Per-channel quantization is more fine-grained, and it uses different $K$ and $B$ for different channels. The latter can achieve higher precision, but it also requires more complicated hardware design. 
\vspace{-10pt}
\subsection{Multipoint Quantization and Optimization}
\label{sec:mpq}

\begin{algorithm}[t]
\caption{Optimization of Problem~\ref{eq:joint}}
\begin{algorithmic}[1]
\label{alg:opt}
\STATE \textbf{Input}: weight $\vec w$, integer $n$, maximal step size for grid search $\eta$, a fixed quantization set $\setvv Q$.
\STATE Initialize the residual $\vec r_1 = \vec w$.
\FOR{$i = 1:n$}
\STATE Compute $\Delta_{\vec r_i}$, the minimal gap of $\vec r_i$, as definition~\ref{def:mingap}.
\STATE Set step size $\gamma$ and search range $I$ as Eq.~\ref{eq:hyper}.
\STATE Solve Eq.~\ref{eq:opta} for \small{$a_i^*$} by grid search in $I$ with step size \small{$\gamma$}.
\STATE Set $\vec w_i^* = \left [ \frac{\vec r_i}{a_i^*} \right ]_{\setvv Q}$, $\vec r_{i+1} = \vec r_i - a_i^* \vec w_i^*$.
\ENDFOR
\STATE Return $\left \{a_i^*, \tilde{\vec w}_i^* \right\}_{i=1}^n$.
\end{algorithmic}
\end{algorithm}

We propose \emph{multipoint quantization}, which can be implemented with common operands on commodity hardware. 


Consider a linear layer in a neural network, which is either a fully-connected (FC) layer or a convolutional layer. The weight of a channel is a vector for FC layer, or a convolution kernel for convolutional layer. For simplicity, we only introduce the case of FC layer in this section. It can be easily generalized to convolutional layers. Supposing the input to this layer is $d$-dimensional
, then the real-valued weight of a channel can be denoted as $\vec w = (w_1, w_2, \dots, w_d)\in \RR^d$. Multipoint quantization approximates $\vv w$ with a weighted sum of  a set of  low precision weight vectors, 
\begin{equation}
    \tilde{\vec w} = \sum_{i=1}^n a_i \tilde{\vec w_i},
\end{equation}
where $a_i \in \R$ and $\tilde{\vec w_i} \in \setvv Q^{d}$ for $\forall i=1,\ldots, n.$ 
Multipoint quantization allows more freedom in representing the full-precision weight. 
Naive quantization approximates a weight by the nearest grid points, while multipoint quantization approximates it with the nearest point on the linear segments. If we release the constraint $a_1 + a_2 = 1$, we can actually represent every point on the 2-dimensional planar with multipoint quantization.
%

%
Given a fixed $n$, we want to find optimal $\{a_i^*, \tilde{\vec w_i}^*\}_{i=1}^n$ that minimizes the $\ell_2$-norm between the real-valued weight  and the weighted sum,
\begin{equation}
\label{eq:joint}
    \{a_i^*, \tilde{\vec w_i}^*\}_{i=1}^n = \argmin_{\{a_i, \tilde{\vec w_i}\}_{i=1}^n}\left|\left|\vec w -  \sum_{i=1}^n a_i \tilde{\vec w_i}\right|\right|.
\end{equation}

Problem~\ref{eq:joint} yields a difficult combinatorial optimization. We are able to get exact approximation when $n=d+1$ by taking $a_i = w_i$ and $\tilde{\vec w_i} = one\_hot(i)$, where $one\_hot(i)$ is a one hot vector with the $i$-th element as 1 and other elements as 0. However, $d$ is always large in deep neural networks, and our goal is to approximate $\vec w$ with a small enough $n$. 
Hence, we propose  an efficient greedy method for solving it, which sequentially adds the best pairs $(a_i, \tilde{\vec w}_i)$ one by one.
Specifically, we obtain the $i$-th pair $(a_i, \tilde{\vec w}_i)$
by approximate the residual from the previous pairs,
\begin{equation}
    (a_i^*, \tilde{\vec w}_i^*) = \argmin_{a,~\tilde{\vec w}}\left|\left| \vec r_i - a \tilde{\vec w} \right|\right|
\end{equation}
where $\vec r_i$ is the residual from the first $i-1$ pairs, 
\begin{equation}
\label{eq:optr}
\vec r_1 = \vec w;~~\vec r_i = \vec w - \sum_{j=1}^{i-1} a_j^* \tilde{\vec w}_j^*,~~~\forall ~i=2,\dots,n   
\end{equation}
For a fixed $a$, 
we have,
\begin{equation}
\begin{aligned}
    \tilde{\vec w}_i^*(a) &= \argmin_{\tilde{\vec w}}\left|\left| \vec r_i - a \tilde{\vec w} \right|\right| \\&= \argmin_{\tilde{\vec w}} \left|\left| \frac{\vec r_i}{a} - \tilde{\vec w} \right|\right| = \left[\frac{\vec r_i}{a} \right]_{\setvv Q}.
\end{aligned}
\end{equation}
Now we only need to solve optimal $a$, 
\begin{equation}
\label{eq:opta}
    a_i^* = \argmin_{a} \left|\left|\vec r_i - a\left[\frac{\vec r_i}{a}\right]_{\setvv Q}\right|\right|.
\end{equation}
Because $[\cdot]_{\setvv Q}$ is not differentiable, it is hard to optimize $a$ by gradient descent. 
Instead, we adopt grid search to find $a_i^*$ efficiently. 
Once the optimal $a_i^*$ is found, 
the corresponding $\tilde{\vec w}_i^*$ is,
\begin{equation}
\label{eq:optw}
    \tilde{\vec w}_i^* = \left[\frac{\vec r_i}{a_i^*}\right]_{\setvv Q}.
\end{equation}

\paragraph{Choice of Parameters for Grid Searching $a_i^*$:} 
Grid search enumerates all the values from set $\left[I_{min}:\gamma:I_{max} \right]$, and selects the value that achieves the lowest error. The parameters of grid search, search range and step size, are defined as the interval $I=[I_{min}, I_{max}]$ and the increment $\gamma$ respectively.
The choice of 
search range $I$ and step size $\gamma$ are critical. 
We first define minimal gap of vector 
, and then give the choice of search range and step size.

The minimal gap is the minimal distance between two elements in a vector $\vec t$. It restricts the maximal value of step size. 
\begin{definition}
\label{def:mingap}
Given vector $\vec t=(t_1,\dots,t_d)\in\R^{d}$,
the \emph{minimal gap} of $\vec t$ is, 
\begin{equation*}
\begin{gathered}
    \Delta_{\vec t} = \min_{i, j} \frac{\left|\,\left|t_{i}\right|-\left|t_{j}\right| \,\right|}{2},\\
    \text{s.t.}~~i, j \in \{1,2,\dots,d\}~~\text{and}~~t_{i} \neq t_{j}.
\end{gathered}
\end{equation*}
\end{definition}

Then we propose the following choice of $I$ and $\gamma$,
\begin{equation}
\label{eq:hyper}
I = [0~,~2 (2^{b-1}-1)||\r_i||];~~\gamma = \min(\frac{\Delta_{\r_{i}}}{2^{b-1}-1},\eta),
\end{equation}

where $\eta$ is a predefined maximal step size to accelerate convergence. In Sec.~\ref{theory}, we show that by choosing $I$ and $\gamma$ like this, our algorithm is guaranteed to converge to zero. As $n$ increases, the dimension of the approximation set increases. Intuitively, the nearest distance from an arbitrary point to the approximation set decreases exponentially with $n$. We rigorously prove that the greedy algorithm decays in an exponential rate in Sec.~\ref{theory}. Algorithm~\ref{alg:opt} recaptures the optimization procedure.

\vspace{-8pt}
\subsection{Multipoint Quantization on Deep Networks}\label{sec:deep}


\begin{algorithm}[t]
\caption{Generate QNN with Multipoint Quantization}
\begin{algorithmic}[1]
\STATE \textbf{Input}: A full-precision network $f$, a predefined threshold $\epsilon$, a calibration set of data points $D = \{\vec x \datai\}_{i=1}^N$.
\STATE Run forward pass of $f$ with calibration set $D = \{\vec x \datai\}_{i=1}^N$ to get the input batch $D_L = \{\vec x_L\datai\}_{i=1}^N$ for each layer $L$ in $f$, 
\FOR{each layer $L$ in $f$}
\FOR{each channel $k$ in layer $L$}
\STATE $\tilde{\vec w_k} \leftarrow \left[\vec w_k \right]_{\setvv Q}$.
\IF{$e(\vec w_k, \tilde{\vec w_k}, D_L) > \epsilon$}
\STATE Apply multipoint quantization with Algorithm~\ref{alg:opt} and keep increasing $n$ until $e(\vec w_k, \tilde{\vec w_k}, D_L) < \epsilon$. Get $\{a_i^*, \tilde{\vec w_i}^*\}_{i=1}^n$.
\STATE $\tilde{\vec w_k} \leftarrow \sum_{i=1}^n a_i^* \tilde{\vec w_i}^*$ 
\ENDIF
\ENDFOR
\ENDFOR
\STATE Return QNN $\tilde{f}$
\end{algorithmic}
\end{algorithm}

We describe how to apply multipoint quantization to deep neural networks. Using multipoint quantization can decrease the quantization error of a channel significantly, but every additional quantized filter requires additional memory and computation consumption. Therefore, to apply it to deep networks, we must select the important channels to compensate for their quantization error with multipoint quantization.


For a layer $L$ with $d$-dimensional input, we adopt a simple criterion, output error, to determine the target channels. Output error is the difference of the output of a channel before and after quantization. Suppose the weight of a channel is $\vec w$, its output error is defined as,
\begin{equation}
    e(\vec w, \vec \tilde{\vec w}, D_L) = \E_{\vec x \sim D_L}||\vec w^\top \vec x - \tilde{\vec w}^\top \vec x||^2_2,
\end{equation}
where $D_L$ is the input batch to $L$, collected by running forward pass of $f$ with calibration set $D$. Our goal is to keep the output of each channel invariant. If $e(\vec w, \tilde{\vec w}, D_L)$ is larger than a predefined threshold $\epsilon$, we apply multipoint quantization to this channel and increase $n$ until $e(\vec w, \tilde{\vec w}, D_L) < \epsilon$.
A similar idea is leveraged to determine the optimal clipping factor $K^*$,
\begin{equation}
\label{eq:clip}
    K^* = \argmin_K \sum_{\vec w \in \setvv W}\E_{\vec x \sim D_L}||\vec w^\top \vec x - \tilde{\vec w}^\top \vec x||^2_2.
\end{equation}
Here, $\setvv W$ is the set of weights sharing the same $K$. For per-layer quantization, $\setvv W$ is contains the weights of all the channels in a layer. For per-channel quantization, $\setvv W$ contains only one element, which is the weight of a channel.

\subsection{Analysis of Overhead}\label{sec:overhead}

We introduce how the computation of dot product can be implemented with common operands when adopting multipoint quantization. Then we analyze the overhead of memory and computation.  

For $d-$dimensional input and weight with $N$ bits, computing the dot product requires $d$ multiplications between two $N-$bit integers. The result of the dot product is stored in a 32-bit accumulator, since the sum of the individual products could be more than $N$ bits. The above operation is called \textit{Multiply-Accumulate (MAC)}, which has been highly optimized in modern deep learning hardware~\citep{chen2016eyeriss}. The 32-bit integer is then quantized according to the quantization scheme of the output part. 

Now we delve into the computation pipeline when $\tilde{\vec w} = \sum_{i=1}^n a_i \tilde{\vec w_i}$. Because $a_i \in \R$, we transform them to a hardware-friendly integer representation beforehand, 
\begin{equation}
\label{eq:quanta}
    a_i \approx \frac{A_i}{2^p}, A_i = [2^p \times a_i]
\end{equation}
Here, $p$ determines the precision of the quantized $a_i$. We use the same $p$ for all the weights with multipoint quantization in the network. $A_i$ are 32-bit integers. The quantization of $a_i$ can be performed off-line before deploying the QNN. We point out that,
\begin{equation}
\begin{aligned}
    \tilde{\vec w}^\top \vec x &= a_1 \tilde{\vec w}_1^\top \vec x + \dots +a_n \tilde{\vec w}_n^\top \vec x\\
    & \approx \frac{A_1 \tilde{\vec w}_1^\top \vec x + \dots + A_n \tilde{\vec w}_n^\top \vec x}{2^p}
\end{aligned}
\end{equation}
We divide the computation into three steps. Readers can refer to Fig.\ref{fig:hardware} in the Appendix for the computational flow charts.

\textbf{Step 1: Matrix Multiplication}~~~~In the first step, we compute $(\tilde{\vec w_1}^\top \vec x, \dots,  \tilde{\vec w_n}^\top \vec x)$. The results are stored in the 32-bit accumulators.

\textbf{Step 2: Coefficient Multiplication \& Summation}~~~~The second step first multiplies $A_i$ with $\tilde{\vec{w}}_i^\top \vec x$, containing $n$ times of multiplication between two 32-bit integers. Then we sum $A_i \tilde{\vec w}_i^\top \vec x$ together with $n-1$ times of addition. 

\textbf{Step 3: Bit Shift} Finally, the division with $2^p$ can be efficiently implemented by shifting $p$ bits of $\sum_{i=1}^n A_i \tilde{\vec w}_i^\top \vec x$ to the left. We ignore the computation overhead in this step.

\begin{table}[!t]
\renewcommand\arraystretch{1.5}
    \centering
    \small{
    \begin{tabular}{cccc}
    \hlinewd{1.2pt}
        Method & Memory & MULs & ADDs \\ \hlinewd{1.2pt}
        Naive & $dN$ & $dN^2$ &$(d-1)N$ \\ \hline
        Multipoint &\small{$ndN+32n$} &\small{$n(dN^2+32^2)$}  & \thead{$n(d-1)N+$ \\$ 32(n-1)$}\\ \hline 
    \end{tabular}}
    \vspace{-10pt}
    \caption{Comparison of memory and computation consumption between a naively quantized layer and a layer using multipoint quantization.}
    \vspace{-10pt}
    \label{tab:overhead}
\end{table}

\textbf{Overall Storage/ Computation Overhead:} We count the number of binary operations following the same bit-op computation strategy as~\citet{li2019additive, zhou2016dorefa}. The multiplication between two $N$-bit integer costs $N^2$ binary operations. Suppose we have a weight vector $\vec w \in \RR^d$. We compare the memory cost and the computational cost (dot product with $N-$bit input $\vec x$) between naive quantization $\tilde{\vec w} = [\vec w]_{\setvv Q}$ and multipoint quantization $\sum_{i=1}^n a_i \tilde{\vec w_i}$. The results are summarized in Table~\ref{tab:overhead}. Because $d$ is always large in neural networks, so the memory and the computation overhead is approximately proportional to the number $n$.

 
\vspace{-5pt}
\section{Theoretical Analysis}
\label{theory}
\begin{figure}[!t]
\centering
\renewcommand{\tabcolsep}{2pt}
\renewcommand\arraystretch{1.1}
\begin{tabular}{c}
\raisebox{4.5em}{\rotatebox{90}{\small{$\log \ell$}}}\includegraphics[width=0.4\textwidth]{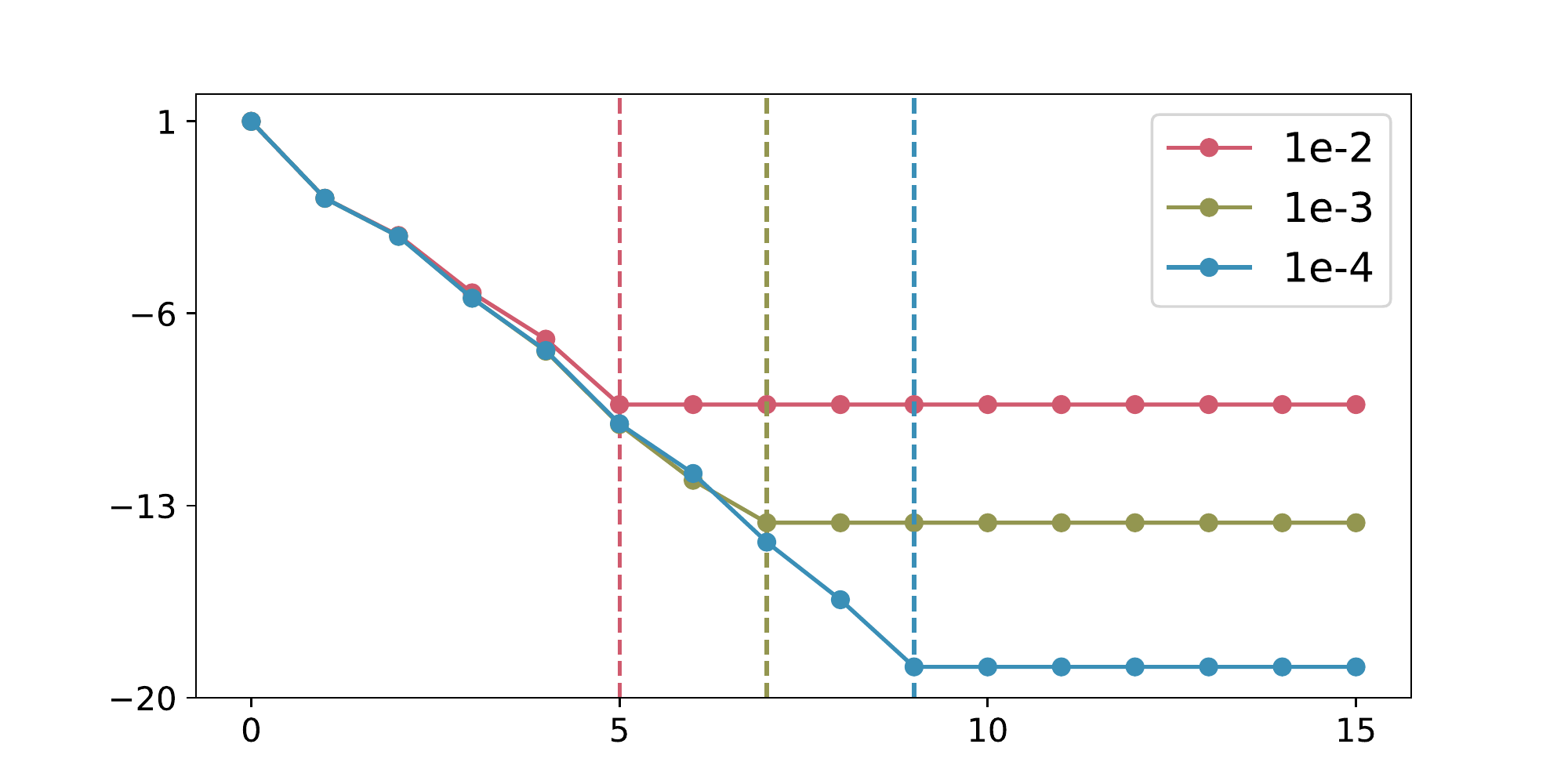}\\
\hspace{1.0em}
{\small{iteration $i$}}
\end{tabular}
\vspace{-8pt}
\caption{A toy experiment where $\vec w$ is randomly generated. The y-axis refers to $\log \ell$. We test three different step sizes $\gamma$ for grid search. As $\ell$ gets smaller, $||\vec r||$ approaches zero and $\Delta_{\vec r}$ also becomes smaller. The dashed line indicates the step that $\gamma>\Delta_{\vec r}$ for the first time. Before the dashed line, $\ell$ decays exponentially. After the dashed line, the grid search does not have enough precision and thus the residual can no more be further reduced.} \label{fig:toy_example}
\vspace{-15pt}
\end{figure}

In this section, we give a convergence analysis of the proposed optimization
procedure. We prove the quantizataion error of the 
 proposed greedy optimization 
 decays 
 exponentially 
w.r.t. the number of points. 

Suppose that we want to quantize
a real-valued $d$-dimensional weight $\w\in\R^{d}$. For simplicity, we assume a binary precision $b=2$ in this section, which leads to $\setvv Q = [-1, 0, 1]$. Our proof can be generalized to $b > 2$ easily. We follow the notations in Section~\ref{sec:mpq}. At the $i$-th iteration, the residual $\vec r_i$, $a_i^*$ and $\tilde{\vec w}_i^*$ are defined by Eq.~\eqref{eq:optr}, Eq.~\eqref{eq:opta} and Eq.~\eqref{eq:optw}, respectively. The minimal gap of a vector $\vec t$, $\Delta_{\vec t}$, is defined in Definition~\eqref{def:mingap}. 

Let the loss function be $\ell(\vec r, a, \vec w) = \left|\left| \vec r - a \vec w \right|\right|$.
Now we can prove the following rate under mild assumptions.
\vspace{-5pt}
\begin{theorem}[Exponential Decay] \label{thm:rate}
Suppose that at the $i$-th iteration of the algorithm,
$a_{i}^*$ is obtained by grid searching $a$ from the range $(0:\gamma_{i}:2 (2^{2-1}-1)||\vec r_i||]$,
where $\gamma_{i}$ is the step size of the grid search. Assume
that $\gamma_{i}\le\min(\Delta_{\r_{i}},\eta)$
for any step $i$ before termination, where $\eta$ is a predefined maximal step size. We have 
\[
\ell(\vec r_i, a_i^*, \vec w_i^*) =\mathcal{O}\left(\exp\left(-ci\right)+\eta\right),
\]
for some constant $c>0$.
\end{theorem}
The proof is in Appendix A. Note that $\eta$ is usually much smaller than the exponential term and thus can be ignored. Theorem 1 suggests that if we use sufficiently small step size ($\gamma_i \le\min(\Delta_{\r_{i}},\eta)$) for the optimization, the loss will decrease exponentially. Because of the exponentially fast decay of the algorithm, we find that $n \leq 2$ for most of the channels using multipoint quantization in practice. Fig.~\ref{fig:toy_example} justifies our theoretical analysis by a toy experiment.

\vspace{-5pt}
\section{Experiments}
We evaluate our method on two tasks, ImageNet classification ~\citep{krizhevsky2012imagenet} and PASCAL VOC object detection~\citep{pascal-voc-2007}. Our evaluation contains various neural networks.
\begin{table*}[!t]
\centering
\renewcommand\arraystretch{1.1}
\small{
\begin{tabular}{c|c|ccccc}
Model &Bits~(W/A)       & Method        & Acc~(Top-1/Top-5) (\%)       & Size & OPs \\ \hlinewd{1.2pt}
\multirow{4}{*}{VGG19-BN} &32/32 &Full-Precision &74.24/91.85 & 76.42MB & -  \\ \cline{2-6}
                           &\multirow{3}{*}{4/8}   & w/o Multipoint & 60.81/83.68 &9.55MB & 9.754G           \\
                                                   && OCS~\small{\citep{zhao2019improving}} & 62.11/84.59 &10.70MB & 10.924G           \\
                                                   && \textbf{Ours}    & \textbf{64.06/86.14} & \textbf{9.59MB} &\textbf{10.923G}   \\ \hlinewd{1.2pt}
\multirow{4}{*}{ResNet-18} &32/32 &Full-Precision &69.76/89.08 & 42.56MB & - \\ \cline{2-6}
                           &\multirow{3}{*}{4/8}   & w/o Multipoint & 54.04/78.10 &5.32MB & 847.78M           \\
                                                   && OCS~\small{\citep{zhao2019improving}} & 58.05/81.57 &6.20MB & 988.51M       \\
                                                   && \textbf{Ours}    & \textbf{61.68/84.03} & \textbf{5.37MB} &\textbf{983.22M}    \\ \hlinewd{1.2pt}
                       
\multirow{4}{*}{ResNet-101} &32/32 &Full-Precision &77.37/93.56 & 161.68MB & - \\ \cline{2-6}
                           &\multirow{3}{*}{4/8}   & w/o Multipoint & 61.04/83.02 &20.21MB & 3.841G           \\
                                                   && OCS~\small{\citep{zhao2019improving}} & 70.27/89.73 &23.40MB & 4.448G           \\
                                                   && \textbf{Ours}    & \textbf{73.09/91.34} & \textbf{20.86MB} &\textbf{4.446G}   \\ \hlinewd{1.2pt}

\multirow{4}{*}{WideResNet-50} &32/32 &Full-Precision &78.51/94.09 & 262.64MB & - \\ \cline{2-6}
                           &\multirow{3}{*}{4/8}   & w/o Multipoint & 61.78/83.60 &31.83MB & 5.639G           \\
                                                   && OCS~\small{\citep{zhao2019improving}} & 68.54/88.68 &35.97MB & 6.372G           \\
                                                   && \textbf{Ours}    & \textbf{70.47/89.43} & \textbf{32.08MB} &\textbf{6.365G}   \\ \hlinewd{1.2pt}
                                                   
\multirow{4}{*}{Inception-v3} &32/32 &Full-Precision &77.45/93.56 & 82.96MB & - \\ \cline{2-6}
                           &\multirow{3}{*}{4/8}   & w/o Multipoint & 5.17/12.85 &10.37MB & 2.846G       \\
                                                   && OCS~\small{\citep{zhao2019improving}} & 8.49/17.75 &12.16MB & 3.338G           \\
                                                   && \textbf{Ours}    & \textbf{33.89/56.07} & \textbf{10.42MB} &\textbf{3.337G}    \\ \hlinewd{1.2pt}

\multirow{4}{*}{Mobilenet-v2} &32/32 &Full-Precision &71.78/90.19 & 8.36MB & - \\ \cline{2-6}
                           &\multirow{3}{*}{8/8}   & w/o Multipoint & 0.06/0.15 &2.090MB & 299.49M           \\
                                                   && OCS~\small{\citep{zhao2019improving}} & N/A &N/A & N/A           \\
                                                   && \textbf{Ours}   & \textbf{70.70/89.70} & \textbf{2.091MB} &\textbf{357.29M}   \\ \hlinewd{1.2pt}
\end{tabular}}
\caption{Per-layer quantization on ImageNet Benchmark (W=Weight, A=Activation, M=$10^6$, G=$10^9$, Acc=Accuracy). \textbf{Bold} refers to the method with highest Top-1 accuracy. Note that OCS cannot be applied to MobileNet-V2 because it cannot deal with group convolution.}
\label{tab:perlayer}
\vspace{-10pt}
\end{table*}

\begin{table*}[!t]
\centering
\renewcommand\arraystretch{1.1}
\small{
\begin{tabular}{c|c|ccccc}
Model &Bits~(W/A)       & Method        & Acc~(Top-1/Top-5) (\%)        & Size & OPs \\ \hlinewd{1.2pt}
\multirow{5}{*}{VGG19-BN} &32/32 &Full-Precision &74.24/91.85 & 76.42MB & -  \\ \cline{2-6}
                           &\multirow{4}{*}{4/4}   & w/o Multipoint & 52.08/76.19 &9.55MB & 4.877G           \\
                                                   && MP~\small{\citep{banner2019post}} & 70.59/90.08 & 9.55MB & 4.877G\\
                                                   && Ours    & 71.96/90.75 & 9.63MB &5.525G   \\
                                                   && \textbf{Ours~+~Clip}    & \textbf{72.78/91.23} & \textbf{9.58MB} &\textbf{5.354G}   \\
                                                   \hlinewd{1.2pt}
\multirow{5}{*}{ResNet-18} &32/32 &Full-Precision &69.76/89.08 & 42.56MB & -  \\ \cline{2-6}
                           &\multirow{4}{*}{4/4}   & w/o Multipoint & 57.00/80.40 &5.32MB & 423.89M\\
                                                   && MP~\small{\citep{banner2019post}} & 64.78/85.90 & 5.32MB & 423.89M\\
                                                   && Ours    & 64.29/85.59 & 5.39MB &494.16M    \\
                                                   && \textbf{Ours~+~Clip} & \textbf{65.89/86.68} & \textbf{5.41MB} &\textbf{470.89M}   \\
                                                   \hlinewd{1.2pt}
\multirow{5}{*}{ResNet-50} &32/32 &Full-Precision &76.15/92.87 & 89.44MB & -  \\ \cline{2-6}
                           &\multirow{4}{*}{4/4}   & w/o Multipoint & 65.88/86.93 &11.18MB & 992.28M \\
                                                   && MP~\small{\citep{banner2019post}} & 72.52/90.80 & 11.18MB & 992.28M\\
                                                   && Ours    & 71.88/90.43 & 11.33MB &1.148G  \\
                                                   && \textbf{Ours~+~Clip}    & \textbf{72.67/91.11} & \textbf{11.32MB} &\textbf{1.128G}   \\
                                                   \hlinewd{1.2pt}
                    
\multirow{5}{*}{ResNet-101} &32/32 &Full-Precision &77.37/93.56 & 161.68MB & - \\ \cline{2-6}
                           &\multirow{4}{*}{4/4}   & w/o Multipoint & 69.67/89.21 &20.21MB & 1.920G           \\
                                                   && \textbf{MP~\small{\citep{banner2019post}}} & \textbf{74.22/91.95} & \textbf{20.21MB} & \textbf{1.920G}\\
                                                   && Ours    & 71.56/90.36 & 20.82MB &2.177G   \\
                                                   && Ours+Clip    & 72.85/91.16 & 21.04MB &2.189G  \\ \hlinewd{1.2pt}
                                                   
\multirow{5}{*}{Inception-v3} &32/32 &Full-Precision &77.45/93.56 & 82.96MB & - \\ \cline{2-6}
                           &\multirow{4}{*}{4/4}   & w/o Multipoint & 12.12/25.24 &10.37MB & 1.423G \\
                                                   
                                                   && MP~\small{\citep{banner2019post}} & 60.64/82.15 & 10.37MB & 1.423G \\
                                                   && Ours    & 61.22/83.27 & 10.44MB &1.692G \\
                                                   && \textbf{Ours+Clip}    & \textbf{65.49/86.72} & \textbf{10.38MB} &\textbf{1.519G}   \\ \hlinewd{1.2pt}
\multirow{5}{*}{Mobilenet-v2} &32/32 &Full-Precision &71.78/90.19 & 8.36MB & - \\ \cline{2-6}
                           &\multirow{4}{*}{4/4}   & w/o Multipoint & 6.86/16.76 &1.04MB & 74.87M \\
                                                   
                                                   && MP~\small{\citep{banner2019post}} & 42.61/67.78 & 1.04MB & 74.87M \\
                                                   && Ours    & 27.52/50.80 & 1.05MB &91.16M \\
                                                   && \textbf{Ours+Clip}    & \textbf{55.54/79.10} & \textbf{1.045MB} &\textbf{85.88M}   \\ \hlinewd{1.2pt}
\end{tabular}}
\caption{Per-channel quantization on ImageNet Benchmark (W=Weight, A=Activation, M=$10^6$, G=$10^9$, MP=Mixed Precision, Acc=Accuracy). Note that MP requires specialized hardware.  \textbf{Bold} refers to the method with highest Top-1 accuracy. `Clip' means using the optimal clipping factor $K^*$ by solving Eq.~\ref{eq:clip}.}
\vspace{-10pt}
\label{tab:perchannel}
\end{table*}

\vspace{-5pt}
\subsection{Experiment Results on ImageNet Benchmark}
\vspace{-3pt}

We evaluate our method on the ImageNet classification benchmark. For fair comparison, we use the pretrained models provided by PyTorch~\footnote{\url{https://pytorch.org/}} as others~\citep{zhao2019improving, banner2019post}. 
We take 256 images from the training set as the calibration set.
Calibration set is used to quantize activations and choose the channels to perform multipoint quantization. 
To improve the performance of low-bit activation quantization, 
we pick the optimal clipping factor for activations by minimizing the mean square error~\citep{sung2015resiliency}. 
Like previous works, the weights of the first and the last layer are always quantized to 8-bit~\citep{nahshan2019loss, li2019additive, banner2019post}. 
For all experiments, we set the maximal step size for grid search in Eq.~\ref{eq:opta} to $\eta = \frac{1}{2^{10}}$.

We report both model size and number of operations under different bit-width settings for all the methods. 
The first and the last layer are not counted. We follow the same bit-op computation strategy as~\citet{li2019additive, zhou2016dorefa} to count the number of binary operations. One OP is defined as one multiplication between an 8-bit weight and an 8-bit activation, which takes 64 binary operations. The multiplication between a $m$-bit and a $n$-bit integer is counted as $\frac{mn}{64}$ OPs. 

We provide two categories of results here: per-layer quantization and per-channel quantization. In per-layer quantization, all the channels in a layer exploit the same $K$ and $B$. In per-channel quantization, each channel has its own parameter $K$ and $B$.
For both settings, 
we test six different networks in our experiments, including VGG-19 with BN~\citep{simonyan2014very}, ResNet-18, ResNet-101, WideResNet-50~\citep{he2016deep}, Inception-v3~\citep{szegedy2015going} and MobileNet-v2~\citep{sandler2018mobilenetv2}.

\textbf{Per-layer Quantization}\space\space
For per-layer quantization, 
we compare our method with a state-of-the-art (SOTA) baseline, Outlier Channel Splitting (OCS)~\citep{zhao2019improving}. OCS duplicates the channel with the maximal absolute value and halves it to mitigate the quantization error. 
For fair comparison, 
we choose the best clipping method among four methods for OCS according to their paper~\citep{sung2015resiliency, migacz20178, banner2019post}.  We select the threshold $\epsilon$ such that the OPs of the QNN with multipoint quantization is about 1.15 times of the naive QNN. For fair comparison, we expand the network with OCS until it has similar OPs with the QNN using multipoint quantization. 
The results without multipoint quantization (denoted `w/o Multipoint' in Table.~\ref{tab:perlayer}) serve as another baseline. 
We quantize the activations and the weights to the same precision as the baselines.
Experiment results are presented in Table.~\ref{tab:perlayer}.
It shows that 
our method obtains consistently significant gain on all the models compared with `w/o Multipoint', with little increase on memory overhead. 
Our method also consistently outperforms the performance of OCS under any computational constraint. 
Especially, on ResNet-18, ResNet-101 and Inception-v3, our method surpasses OCS by more than 2\% Top-1 accuracy. 
OCS cannot quantize MobileNet-v2 due to the group convolution layers, while our method nearly recovers the full-precision accuracy. 
Our method achieves similar performance with Data Free Quantization~\citep{nagel2019data} (71.19\% Top-1 accuracy with 8-bit MobileNet-v2), which focuses on 8-bit quantization on MobileNets only. 
Note that this method is orthogonal to ours and
we expect to obtain more improvement by combining with it.

\textbf{Per-channel Quantization}\space\space
For per-channel quantization, 
we compare our method with another SOTA baseline, ~\citet{banner2019post}. \citet{banner2019post} requires physical per-channel mixed precision computation since it assigns different bits to different channels. We denote it as 'Mixed Precision (MP)'. All networks are quantized with asymmetric per-channel quantization ($B \neq 0$). Since per-channel quantization has higher precision, weight clipping is not performed for naive quantization, which means that $K=\max(|\vec w|)$. We quantize both weights and activations to 4 bits. Experiment results are presented in Table.~\ref{tab:perchannel}. 

Our method outperforms MP on VGG19-BN and Inception-v3 even without weight clipping. After performing weight clipping with Eq.~\ref{eq:clip}, our method beats MP on 5 out of 6 networks, except for ResNet-101. 
On VGG19-BN, Inception-v3 and MobileNet-v2, compared with MP, 
the Top-1 accuracy of our method after clipping is more than 2\% higher. 
In the experiments, all the memory overhead is smaller than 5\% and the computation overhead is no more than 17\% compared with the naive QNN. 
\vspace{-5pt}
\subsection{Experiment Results on PASCAL VOC Object Detection Benchmark}
\vspace{-2pt}
We test Single Shot MultiBox Object Detector (SSD), which is a well-known object detection framework. We use an open-source implementation~\footnote{\url{https://github.com/amdegroot/ssd.pytorch}}. The backbone network is VGG16. We apply per-layer quantization and per-channel quantization on all the layers, excluding localization layers and classification layers. Due to the GPU memory constraint, the calibration set only contains 6 images. We measure the mean average precision (mAP), size and OPs of the quantized model. We perform activation clipping and weight clipping for both settings.

In per-layer quantization, our method increases the performance of the baseline by over 1\% mAP (72.86\% $\longrightarrow$ 74.10\%). When weight is quantized to 3-bit, our method boost the baseline by 4.38\% mAP (42.56\% $\longrightarrow$ 46.94\%) with little memory overhead of 0.01MB. Our method also performs well in per-channel quantization. It improves the baseline by 0.41\% mAP for 4-bit quantization and 1.09\% mAP for 3-bit quantization. Generally, our method performs better when the bit width goes smaller.
\vspace{-5pt}
\begin{table}[!t]

\centering
\renewcommand\arraystretch{1.0}
\small{
\subtable[Per-layer Quantization]{
\begin{tabular}{c|c|cccc}
\hlinewd{1.2pt}
W/A       & Method        & mAP(\%)        & Size(MB) & OPs(G) \\ \hlinewd{1.2pt}
32/32 & FP &77.43 & 100.24 & -  \\ \hline
\multirow{2}{*}{4/8}    & w/o Multipoint & 72.86 &12.53 & 15.69 \\
                        & \textbf{Ours}    & \textbf{74.10} & \textbf{12.63} &\textbf{17.58}   \\ \hline
\multirow{2}{*}{3/8}    & w/o Multipoint & 42.56 &9.40 & 11.76 \\
                        & \textbf{Ours}    & \textbf{46.94} & \textbf{9.41} &\textbf{12.18}   \\ \hline
\end{tabular}}

\subtable[Per-channel Quantization]{
\begin{tabular}{c|c|cccc}
\hlinewd{1.2pt}
W/A       & Method        & mAP(\%)       & Size(MB) & OPs(G) \\ \hlinewd{1.2pt}
32/32 &FP &77.43 & 100.24 & -  \\ \hline
\multirow{2}{*}{4/4}    & w/o Multipoint & 73.17 &12.53 & 7.843 \\
                        & \textbf{Ours}    & \textbf{73.58} & \textbf{12.62} &\textbf{8.636}   \\ \hline
\multirow{2}{*}{3/3}    & w/o Multipoint & 59.37 &9.40 & 4.412 \\
                        & \textbf{Ours}    & \textbf{60.46} & \textbf{9.43} &\textbf{4.733}   \\ \hline
\end{tabular}}
}
\vspace{-10pt}
\caption{Post-training quantization result on SSD-VGG16 (mAP=mean average precision, FP=Full-Precision). \textbf{Bold} refers to the method with highest mAP.}
\label{tab:detection}
\vspace{-20pt}
\end{table}

\begin{figure}[!t]
\centering
\renewcommand{\tabcolsep}{2pt}
\renewcommand\arraystretch{0.5}
\begin{tabular}{c}
\raisebox{2.5em}{\rotatebox{90}{\small{Top-1 Accuracy/\%}}}\includegraphics[width=0.4\textwidth]{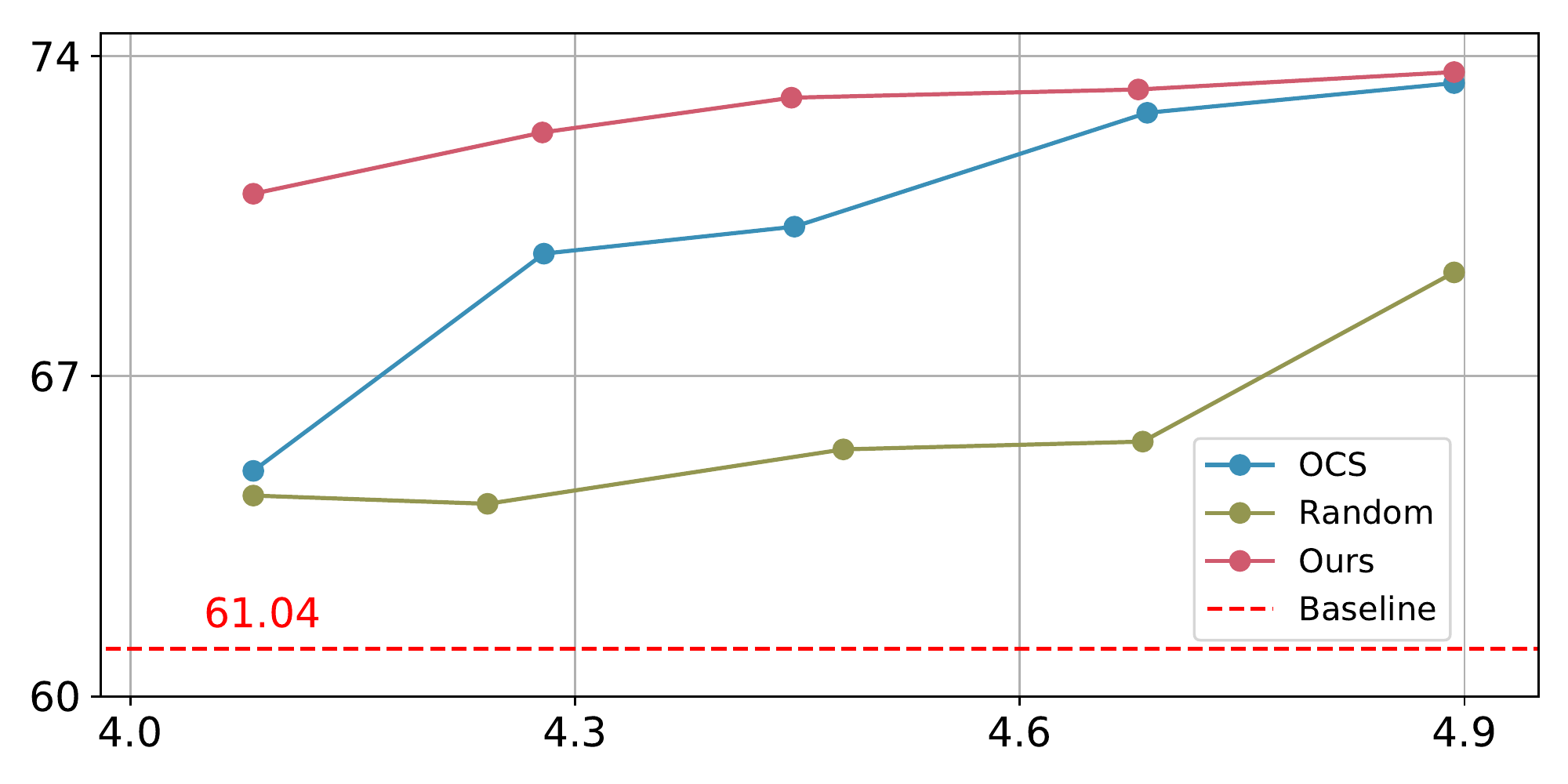}\\
\hspace{1.0em}
{\small{OPs/G}}
\end{tabular}
\vspace{-10pt}
\caption{The trade-off between computational cost and performance of a quantized ResNet-101 (W4A8). `Baseline' is the naive QNN without multipoint quantization (OPs=3.841G). `Random' uses multipoint quantization but channels are randomly added, while our method adds channels according to their output error.} \label{fig:linechart}
\vspace{-13pt}
\end{figure}

\begin{figure}[!t]
\centering
\renewcommand{\tabcolsep}{2pt}
\renewcommand\arraystretch{0.5}
\begin{tabular}{c}
\raisebox{0.8em}{\rotatebox{90}{\small{Relative Increment of Size}}}\includegraphics[width=0.4\textwidth]{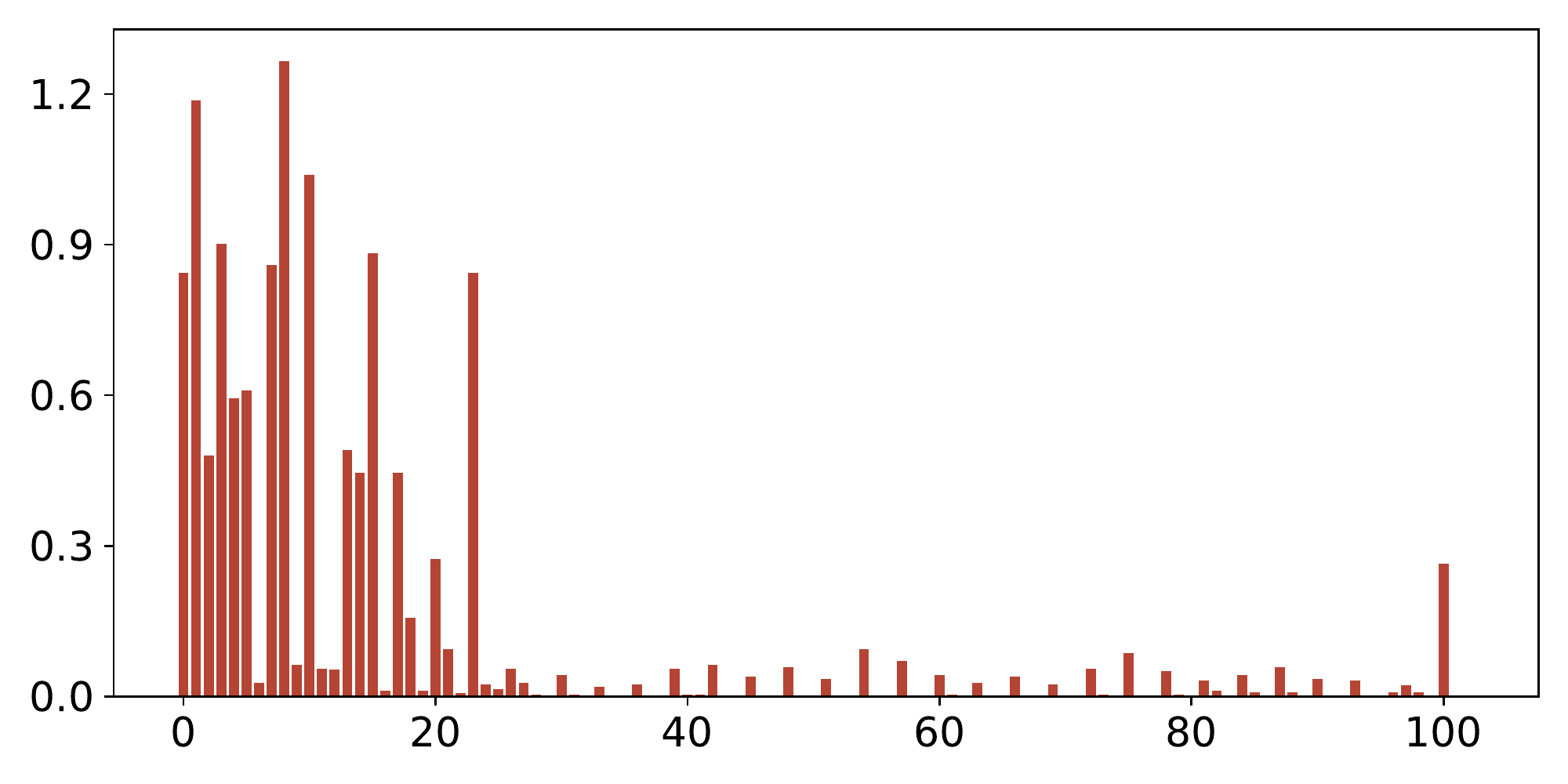}\\
\hspace{1.0em}
{\small{Index of Layers}}
\end{tabular}
\caption{Relative increment of size in each layer of a quantized ResNet-101 with multipoint quantization. The layers close to the input have large relative increment, while the layers after the 30-th layer have only negligible increment.} \label{fig:neuron_increment}
\vspace{-15pt}
\end{figure}
\vspace{-5pt}
\subsection{Analysis of the Algorithm}
\vspace{-3pt}
\label{sec:analysis}
We provide a case study of ResNet-101 under per-layer quantization to analyze the algorithm. More results can be found in the appendix.

\textbf{Computation Overhead and Performance:} 
Fig.~\ref{fig:linechart} demonstrates how the performance of different methods changes as the computational cost changes. Our method obtains huge gain with only a little overhead. OCS cannot perform comparably with our method at the beginning, but it catches up when the computational cost is large enough. The performance of `Random' is consistently the worst among all three methods, implying the importance of choosing appropriate channels for multipoint quantization.       

\textbf{Where Multipoint Quantization is Applied:}
Fig~\ref{fig:neuron_increment} shows the relative increment of size in each layer. We observe that the layers close to the input have more relative increment of size compared with later layers.


\vspace{-5pt}
\section{Conclusions}
We propose multipoint quantization for post-training quantization, which is hardware-friendly and effective. It performs favorably compared with state-of-the-art methods.
\newpage
\bibliography{Main.bib}

\newpage
\appendix
\onecolumn

\section{A. Proof of Theorem \ref{thm:rate}} \label{apx:proof}


Notice that in the main text, we define $a_{i}^{*},\tilde{\w}_{i}^{*}=\underset{a,\tilde{\w}}{\arg\min}\left\Vert \r_{i}-a\tilde{\w}\right\Vert$ as the optimal solution of each iterations (see Equ (6)). While this can not be solved and in practice we use $a_{i}^{*}=\underset{a}{\arg\min}\left\Vert \r_{i}-a\left[\frac{r_{i}}{a}\right]_{\Q}\right\Vert$ and $\tilde{\w}_{i}^{*}=\left[\frac{\r_{i}}{a_{i}^{*}}\right]_{\Q}$. This slightly abuses the notation as the two $a_i^*$ and $\tilde{\w_i^*}$ are actually different. We do this mainly for notation simplicity in the main text. In the proof we distinguish the notations. We use 
$
a_{i}^{*},\tilde{\w}_{i}^{*}=\underset{a,\tilde{\w}}{\arg\min}\left\Vert \r_{i}-a\tilde{\w}\right\Vert,
$
$a_{i}=\underset{a}{\arg\min}\left\Vert \r_{i}-a\left[\frac{r_{i}}{a}\right]_{\Q}\right\Vert$ and $\tilde{\w}_{i}=\left[\frac{\r_{i}}{a_{i}}\right]_{\Q}$ in this proof.

In our proof, we only consider the simplest case when $b=2$, which means $\Q = \{-1, 0, 1\}$. It can be generalized to $b > 2$ easily. Define $\Q_{\text{vec}}:=\left\{ \left[\boldsymbol{v}\right]_{\Q}\mid\boldsymbol{v}\in\R^{d}\right\}$, $\Q^{\w}_{\text{vec}}:=\left\{ 2\left\Vert \w\right\Vert \left[\boldsymbol{v}\right]_{\Q}\mid\boldsymbol{v}\in\R^{d}\right\}$ and $\M = \text{conv}(\Q^{\w}_{\text{vec}})$,
where $\text{conv}(S)$ denotes the convex hull of set $S$. It is
obvious that $\boldsymbol{w}$ is an interior point of $\M$. Now
we define the following intermediate update scheme. Given the current
residual vector $\r_{i}$, without loss of generality, we assume all the elements of $r_i$ are different (if some of them are equal, we can simply treat them as the same elements). We define 
\begin{align*}
\tilde{\w}'_{i} & =\underset{\w\in\M}{\arg\max}\left\langle \r_{i},\w\right\rangle \\
a'_{i} & =\underset{a\in[0,1]}{\arg\min}\left\Vert \r_{i}-a\tilde{\w}'_{i}\right\Vert .
\end{align*}
Notice that as the objective is linear and we thus have 
\[
\tilde{\w}'_{i}=\underset{\w\in\Q^{\w}_{\text{vec}}}{\arg\max}\left\langle \r_{i},\w\right\rangle .
\]
Without loss of generality, we assume $\tilde{\w}'_{i}\neq\boldsymbol{0}$,
as in this case, we have $\boldsymbol{r_{i}}=\boldsymbol{0}$ and
the algorithm should be terminated. Simple algebra shows that $a'_{i}=\frac{\left\langle \tilde{\w}'_{i},\r_{i}\right\rangle }{\left\Vert \tilde{\w}'_{i}\right\Vert ^{2}}$.
Notice that as we assume $\tilde{\w}'_{i}\neq\boldsymbol{0}$, we have $\left\Vert \tilde{\w}'_{i}\right\Vert \ge2\left\Vert \w\right\Vert $.
This gives that
\[
a'_{i}\le\frac{1}{2\left\Vert \w\right\Vert }\left\langle \frac{\tilde{\w}'_{i}}{\left\Vert \tilde{\w}'_{i}\right\Vert },\r_{i}\right\rangle \le\frac{\left\Vert \r_{i}\right\Vert }{2\left\Vert \w\right\Vert }\le\frac{\left\Vert \r_{0}\right\Vert }{2\left\Vert \w\right\Vert }=\frac{1}{2}.
\]
Hence the optimal solution under the constraint of $a'_i \in [0, 1]$ is also $a'_i = \frac{\left\langle \tilde{\w}'_{i},\r_{i}\right\rangle }{\left\Vert \tilde{\w}'_{i}\right\Vert ^{2}}$. Given the current residual vector, we also define 
\[
\left(a_{i}^{*},\tilde{\w}_{i}^{*}\right)=\underset{a\in[0,1],\w\in\Q^{\w}_{\text{vec}}}{\arg\min}\left\Vert \r_{i}-a\w\right\Vert .
\]
By the definition, we have $\left\Vert \r_{i}-a_{i}^{*}\tilde{\w}_{i}^{*}\right\Vert \le\left\Vert \r_{i}-a'_{i}\tilde{\w}'_{i}\right\Vert .$
We have the following inequalities:
\begin{align*}
\left\Vert \r_{i}-a_{i}^{*}\tilde{\w}_{i}^{*}\right\Vert ^{2} & \le\left\Vert \r_{i}-a'_{i}\tilde{\w}'_{i}\right\Vert ^{2}\\
 & =\left\Vert \r_{i}\right\Vert ^{2}-2a'_{i}\left\langle \tilde{\w}'_{i},\r_{i}\right\rangle +(a'_{i})^{2}\left\Vert \tilde{\w}'_{i}\right\Vert ^{2}\\
 & =\left\Vert \r_{i}\right\Vert ^{2}-\left(\frac{\left\langle \tilde{\w}'_{i},\r_{i}\right\rangle }{\left\Vert \tilde{\w}'_{i}\right\Vert }\right)^{2}.
\end{align*}
Notice that as we showed that $\boldsymbol{w}$ is an interior point
of $\M$, we have $\left(\frac{\left\langle \tilde{\w}'_{i},\r_{i}\right\rangle }{\left\Vert \tilde{\w}'_{i}\right\Vert }\right)^{2}\ge q\left\Vert \r_{i}\right\Vert ^{2}$,
for some $q\in(0,1]$. This gives that 
\[
\left\Vert \r_{i}-a_{i}^{*}\tilde{\w}_{i}^{*}\right\Vert ^{2}\le(1-q)\left\Vert \r_{i}\right\Vert ^{2}.
\]

We define 
\[
\left(a_{i}^{**},\tilde{\w}_{i}^{**}\right)=\underset{a\in[0,2\left\Vert \w\right\Vert ],\w\in\Q_{\text{vec}}}{\arg\min}\left\Vert \r_{i}-a\w\right\Vert .
\]
And it is obvious that we have $\left\Vert \r_{i}-a_{i}^{*}\tilde{\w}_{i}^{*}\right\Vert ^{2}=\left\Vert \r_{i}-a_{i}^{**}\tilde{\w}_{i}^{**}\right\Vert ^{2}$.
Next we bound the difference between $\left\Vert \r_{i}-a_{i}^{**}\tilde{\w}_{i}^{**}\right\Vert ^{2}$
and $\left\Vert \r_{i}-a_{i}\tilde{\w}_{i}\right\Vert ^{2}.$ Notice
that for any $a>0$, we have
\[
\left[\frac{q}{a}\right]=1*\mathbb{I}\left\{ q\ge0.5a\right\} +0*\mathbb{I}\left\{ q\in(-0.5a,0.5a)\right\} -1*\mathbb{I}\left\{ q\le-0.5a\right\} .
\]
Without loss of generality we assume that $\left|\left|r_{i,1}\right|-0.5a_{i}^{**}\right|\le\left|\left|r_{i,j}\right|-0.5a_{i}^{**}\right|$,
for any $j\ge2$. Without loss of generality, we also suppose that
$|r_{i,1}|\ge0.5a_{i}^{**}$. Under the assumption of grid search, there
exists $a''_{i}$ in the search space such that $a_{i}^{**}-\Delta_{\r_{i}}\le a''_{i}\le a_{i}^{**}$.
For any $j\ge2$, if $\left|r_{i,j}\right|\ge0.5a_{i}^{**}$, then
$\left|r_{i,j}\right|\ge0.5a_{i}^{''}$. Now we consider the case
of $\left|r_{i,j}\right|<0.5a_{i}^{**}$. By the assumption that $\left|\left|r_{i,1}\right|-0.5a_{i}^{**}\right|\le\left|\left|r_{i,j}\right|-0.5a_{i}^{**}\right|$,
for any $j\ge2$, we have 
\[
0.5a_{i}^{**}-\left|r_{i,j}\right|\ge\left|r_{i,1}\right|-0.5a_{i}^{**}\Longrightarrow a_{i}^{**}\ge\left|r_{i,1}\right|+\left|r_{i,j}\right|.
\]
This gives that
\begin{align*}
 & 0.5a''_{i}-\left|r_{i,j}\right|\\
\ge & 0.5\left(a_{i}^{**}-\Delta_{\r_{i}}\right)-\left|r_{i,j}\right|\\
\ge & 0.5\left(\left|r_{i,1}\right|+\left|r_{i,j}\right|-\Delta_{\r_{i}}\right)-\left|r_{i,j}\right|\\
= & 0.5\left(\left|r_{i,1}\right|-\left|r_{i,j}\right|-\Delta_{\r_{i}}\right)\\
= & 0.5\left(\left|\left|r_{i,1}\right|-\left|r_{i,j}\right|\right|-\Delta_{\r_{i}}\right)\\
\ge & 0.25\left|\left|r_{i,1}\right|-\left|r_{i,j}\right|\right|\\
> & 0.
\end{align*}
Here the last inequality is from the assumption that $\Delta_{\r_{i}}>0$.
Thus we have for any $j\in\{1,...,d\}$, $\left[\frac{r_{i,j}}{a_{i}^{**}}\right]=\left[\frac{r_{i,j}}{a''_{i}}\right]$.
The case for $r_{i,1}<0.5a_{i}^{**}$ is similar by choosing $a_{i}^{**}\le a''_{i}\le a_{i}^{**}+\Delta_{\r_{i}}$.
This concludes that we have $a''_{i}$ in the search region such that
$\left|a_{i}^{**}-a''_{i}\right|\le\Delta_{\r_{i}}$ and $\left[\frac{\r_{i}}{a_{i}^{**}}\right]_\Q=\left[\frac{\r_{i}}{a''_{i}}\right]_\Q.$
Thus we have
\begin{align*}
 & \left\Vert \r_{i}-a''_{i}\left[\frac{\r_{i}}{a''_{i}}\right]_\Q\right\Vert ^{2}-\left\Vert \r_{i}-a_{i}^{**}\tilde{\w}_{i}^{**}\right\Vert ^{2}\\
= & -2\left(a''_{i}-a_{i}^{**}\right)\left\langle \r_{i},\left[\frac{\r_{i}}{a''_{i}}\right]_\Q\right\rangle +\left(\left(a''_{i}\right)^{2}-\left(a_{i}^{**}\right)^{2}\right)\left\Vert \left[\frac{\r_{i}}{a''_{i}}\right]_\Q\right\Vert ^{2}\\
\le & \eta c,
\end{align*}
for some constant $c$. We have
\begin{align*}
\left\Vert \r_{i}-a_{i}\tilde{\w}_{i}\right\Vert ^{2} & \le\left\Vert \r_{i}-a''_{i}\left[\frac{\r_{i}}{a''_{i}}\right]_\Q\right\Vert ^{2}\\
 & \le\left\Vert \r_{i}-a_{i}^{**}\tilde{\w}_{i}^{**}\right\Vert ^{2}+c\eta\\
 & \le(1-q)\left\Vert \r_{i}\right\Vert ^{2}+c\eta
\end{align*}
This gives that 
\[
\left(\left\Vert \r_{i}-a_{i}\tilde{\w}_{i}\right\Vert ^{2}-\frac{c\eta}{q}\right)\le(1-q)\left(\left\Vert \r_{i}\right\Vert ^{2}-\frac{c\eta}{q}\right).
\]
Apply the above inequality iteratively, we have 
\[
\left\Vert \r_{i}\right\Vert^2 =\mathcal{O}\left((1-q)^{i}+\eta\right).
\]

\section{B. Experiment Details}
We provide more details of our algorithm in the experiments. For per-layer and per-channel quantization, the optimal clipping factor are obtained by uniform grid search from $\left[ 0.05:0.05:1\right] \times \max(|\vec a|)$. For the first and last layer, we search for the optimal clipping factor on weights from $\left[ 0.05:0.05:1\right] \times \max(|\vec w|)$. The optimal clipping factors for weights are obtained before performing multipoint quantization and we keep them fixed afterwards. For fair comparison, the quantization of the Batch Normalization layers are quantized in the same way as the baselines. When comparing with OCS, the BN layers are not quantized. When comparing with ~\citep{banner2019post}, the BN layers are absorbed into the weights and quantized together with the weights. Similar strategy for SSD quantization is adopted, i.e., the BN layers are kept full-precision for per-layer setting and absorbed in the per-channel setting.

The hyperparameter $\epsilon$ for different networks in different settings are listed in Table~\ref{tab:ep_layer} and Table~\ref{tab:ep_chan}.

\begin{table}[htbp]
    \centering
    \begin{tabular}{c|cccccc}
         \hline
         Network & VGG19-BN & ResNet-18 & ResNet-101 & WideResNet-50 & Inception-v3 & Mobilenet-v2 \\ \hline
         $\epsilon$ & 50 & 15 & 0.25 & 1 & 100 & 10 \\
         \hline
    \end{tabular}
    \caption{$\epsilon$ for per-layer quantization (W/A = 4/8)}
    \label{tab:ep_layer}
\end{table}

\begin{table}[htbp]
    \centering
    \begin{tabular}{c|cccccc}
         \hline
         Network & VGG19-BN & ResNet-18 & ResNet-50 & ResNet-101 & Inception-v3 & Mobilenet-v2 \\ \hline
         $\epsilon$ & 10 & 8 & 0.7 & 0.2 & 50 & 1 \\
         \hline
    \end{tabular}
    \caption{$\epsilon$ for per-channel quantization (W/A = 4/4)}
    \label{tab:ep_chan}
\end{table}

\section{C. 3-bit Quantization}
We present the results of 3-bit quantization in this section. 3-bit quantization is more aggressive and the accuracy of the QNN is typically much lower than 4-bit. As before, we report the results of per-layer quantization and per-channel quantization. All the hyper-parameters are the same as 4-bit quantization except for $\epsilon$.

\begin{table*}[htbp]
\centering
\renewcommand\arraystretch{1.2}
\begin{tabular}{c|c|ccccc}
Model &Bits~(W/A)       & Method        & Acc~(Top-1/Top-5)        & Size & OPs & $\epsilon$\\ \hlinewd{1.2pt}
\multirow{3}{*}{VGG19-BN} &32/32 &Full-Precision &74.24\%/91.85\% & 76.42MB & - &- \\ \cline{2-7}
                           &\multirow{2}{*}{3/8}   & w/o Multipoint & 4.71\%/12.33\% &7.16MB & 7.315G & -          \\
                                                   && \textbf{Ours}    & \textbf{20.58\%/40.38\%} & \textbf{7.22MB} &\textbf{8.648G} &\textbf{100}   \\ \hlinewd{1.2pt}
\multirow{3}{*}{ResNet-18} &32/32 &Full-Precision &69.76\%/89.08\% & 42.56MB & - &- \\ \cline{2-7}
                           &\multirow{2}{*}{3/8}   &  w/o Multipoint & 9.83\%/24.89\% &3.99MB & 635.83M & -          \\
                                                   && \textbf{Ours}    & \textbf{26.16\%/49.29\%} & \textbf{4.01MB} &\textbf{714.53M} &\textbf{100}   \\ \hlinewd{1.2pt}

\multirow{3}{*}{WideResNet-50} &32/32 &Full-Precision &78.51\%/94.09\% & 262.64MB & - &- \\ \cline{2-7}
                           &\multirow{2}{*}{3/8}   & No Boosting & 4.36\%/10.64\% &23.87MB & 4.229G & -          \\

                                                   && \textbf{Ours}    & \textbf{18.43\%/35.34\%} & \textbf{23.97MB} &\textbf{4.554G} &\textbf{5}   \\ \hlinewd{1.2pt}
\end{tabular}
\caption{Per-layer quantization on ImageNet Benchmark (W=Weight, A=Activation, M=$10^6$, G=$10^9$, Acc=Accuracy)}
\label{tab:perlayer}
\end{table*}

\begin{table*}[htbp]
\centering
\renewcommand\arraystretch{1.2}
\begin{tabular}{c|c|ccccc}
Model &Bits~(W/A)       & Method        & Acc~(Top-1/Top-5)        & Size & OPs & $\epsilon$\\ \hlinewd{1.2pt}
\multirow{3}{*}{VGG19-BN} &32/32 &Full-Precision &74.24\%/91.85\% & 76.42MB & - &- \\ \cline{2-7}
                           &\multirow{2}{*}{3/3}   & w/o Multipoint & 0.10\%/0.492\% &7.16MB & 2.743G & -          \\
                                                   && \textbf{Ours~+~Clip}    & \textbf{65.81\%/87.25\%} & \textbf{7.19MB} &\textbf{3.099G} &\textbf{50}   \\
                                                   \hlinewd{1.2pt}
\multirow{3}{*}{ResNet-18} &32/32 &Full-Precision &69.76\%/89.08\% & 42.56MB & - &- \\ \cline{2-7}
                           &\multirow{2}{*}{3/3}   & w/o Multipoint & 0.11\%/0.55\% &3.99MB & 238.44M & -          \\
                                                   && \textbf{Ours~+~Clip} & \textbf{43.75\%/69.16\%} & \textbf{4.06MB} &\textbf{265.90M} &\textbf{20}   \\
                                                   \hlinewd{1.2pt}
                                                   
\multirow{3}{*}{MobileNet-v2} &32/32 &Full-Precision &71.78\%/90.19\% & 8.36MB & - &- \\ \cline{2-7}
                           &\multirow{2}{*}{3/3}   & w/o Multipoint & 0.11\%/0.64\% &0.78MB & 42.12M & -          \\
                                                   && \textbf{Ours+Clip}    & \textbf{5.21\%/14.33\%} & \textbf{0.79MB} &\textbf{58.65M} &\textbf{50}   \\ \hlinewd{1.2pt}

\end{tabular}
\caption{Per-channel quantization on ImageNet Benchmark (W=Weight, A=Activation, M=$10^6$, G=$10^9$, Acc=Accuracy)}
\label{tab:perchannel}
\end{table*}

\newpage
\twocolumn
\section{D. Additional Figures}

\begin{figure}[htbp]
\centering
\renewcommand{\tabcolsep}{2pt}
\renewcommand\arraystretch{0.5}
\begin{tabular}{c}
\raisebox{2.5em}{\rotatebox{90}{\small{Top-1 Accuracy/\%}}}\includegraphics[width=0.4\textwidth]{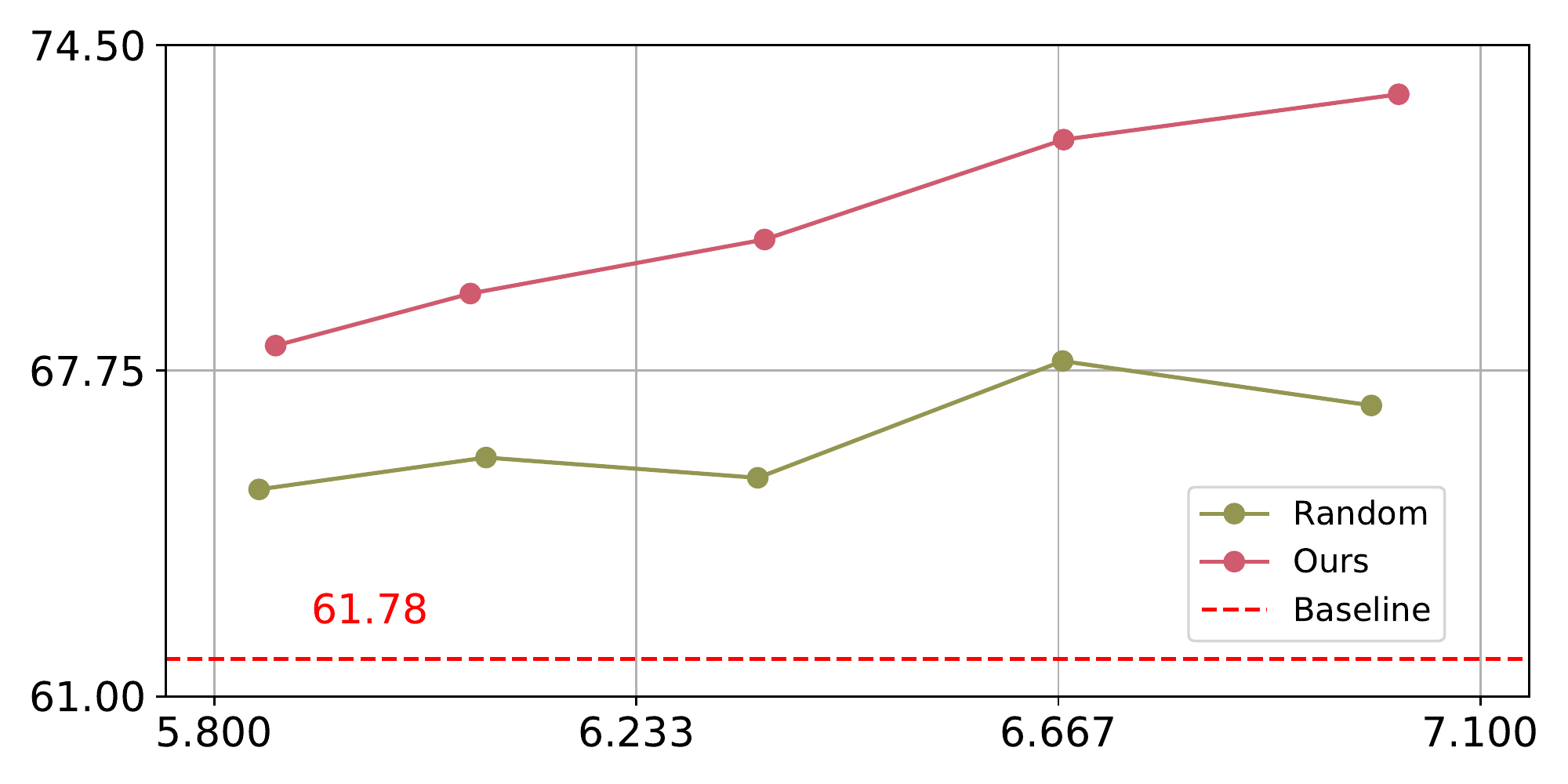}\\
\hspace{1.0em}
{\small{OPs/G}}
\end{tabular}
\caption{The trade-off between computational cost and performance of a per-layer quantized WideResNet-50 (W4A8).} \label{fig:linechart_wideresnet50}
\vspace{-13pt}
\end{figure}

\begin{figure}[htbp]
\centering
\renewcommand{\tabcolsep}{2pt}
\renewcommand\arraystretch{0.5}
\begin{tabular}{c}
\raisebox{0.8em}{\rotatebox{90}{\small{Relative Increment of Size}}}\includegraphics[width=0.4\textwidth]{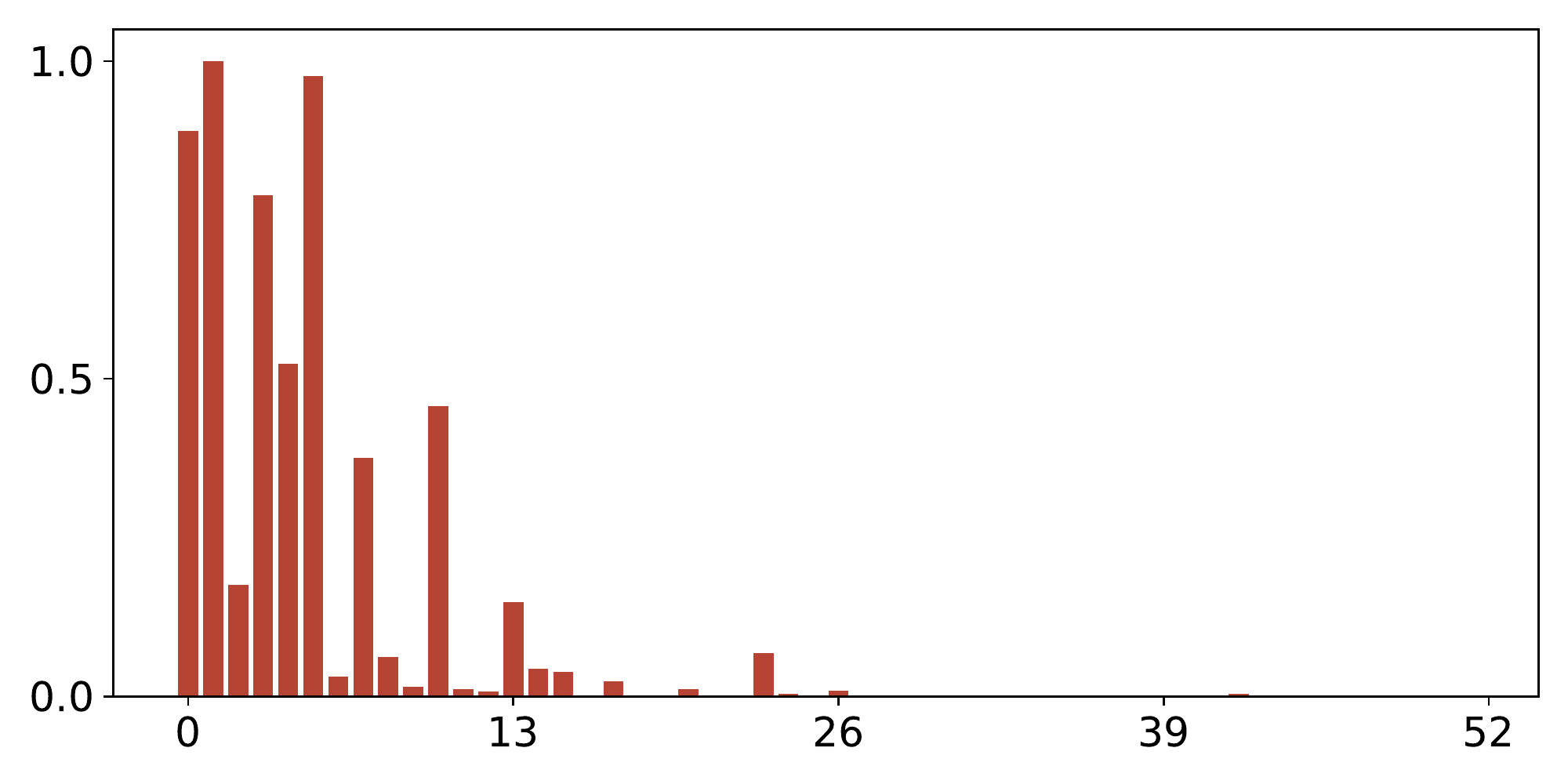}\\
\hspace{1.0em}
{\small{Index of Layers}}
\end{tabular}
\caption{Relative increment of size in each layer of a per-layer quantized WideResNet-50 with multipoint quantization (W4A8).} \label{fig:neuron_increment_wideresnet50}
\vspace{-15pt}
\end{figure}

\begin{figure}[htbp]
\centering
\renewcommand{\tabcolsep}{2pt}
\renewcommand\arraystretch{0.5}
\begin{tabular}{c}
\raisebox{2.5em}{\rotatebox{90}{\small{Top-1 Accuracy/\%}}}\includegraphics[width=0.4\textwidth]{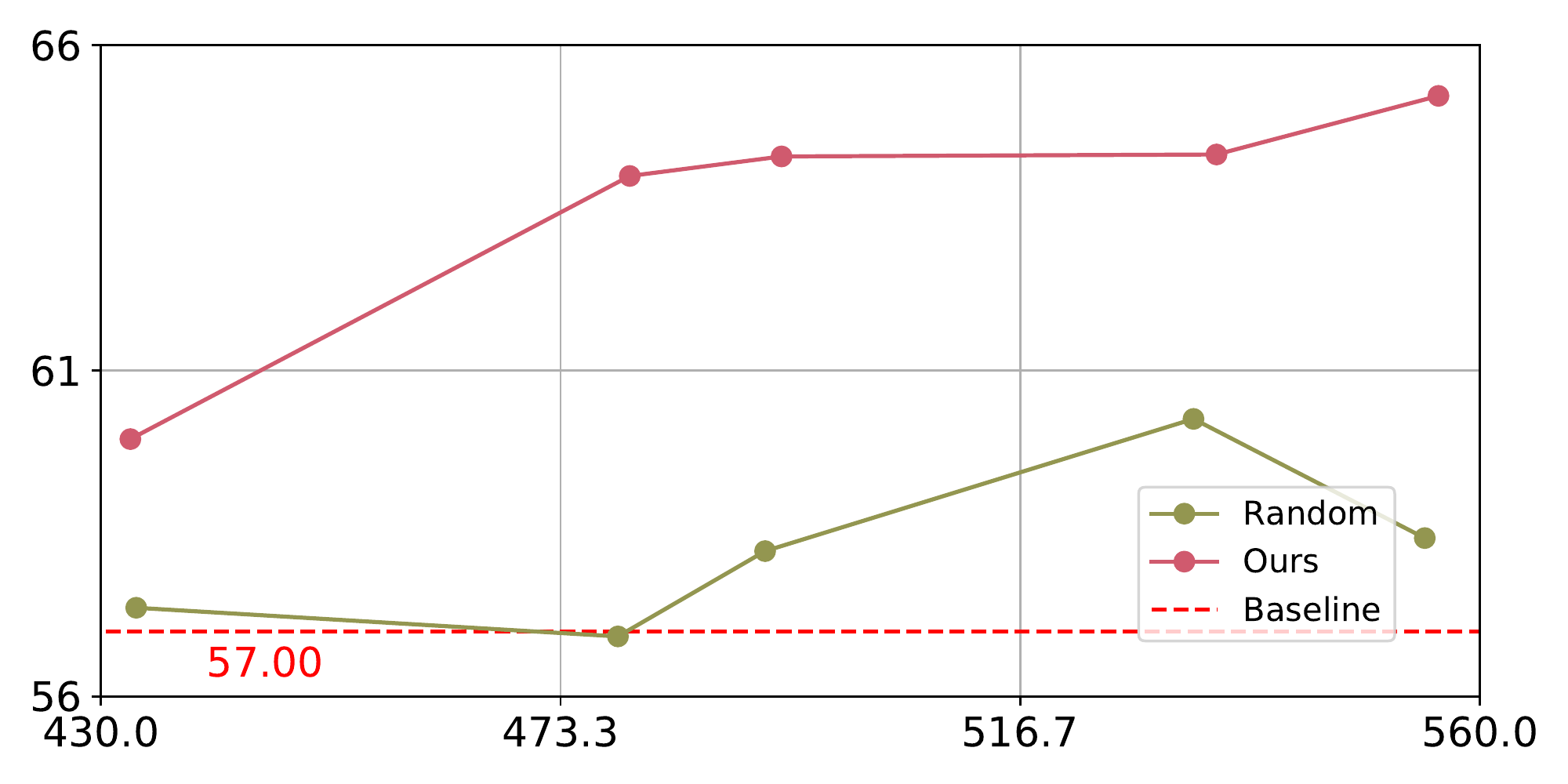}\\
\hspace{1.0em}
{\small{OPs/M}}
\end{tabular}
\caption{The trade-off between computational cost and performance of a per-channel quantized ResNet-18 (W4A4).} \label{fig:linechart_resnet18}
\vspace{-13pt}
\end{figure}

\begin{figure}[!b]
\centering
\renewcommand{\tabcolsep}{2pt}
\renewcommand\arraystretch{0.5}
\begin{tabular}{c}
\raisebox{0.8em}{\rotatebox{90}{\small{Relative Increment of Size}}}\includegraphics[width=0.4\textwidth]{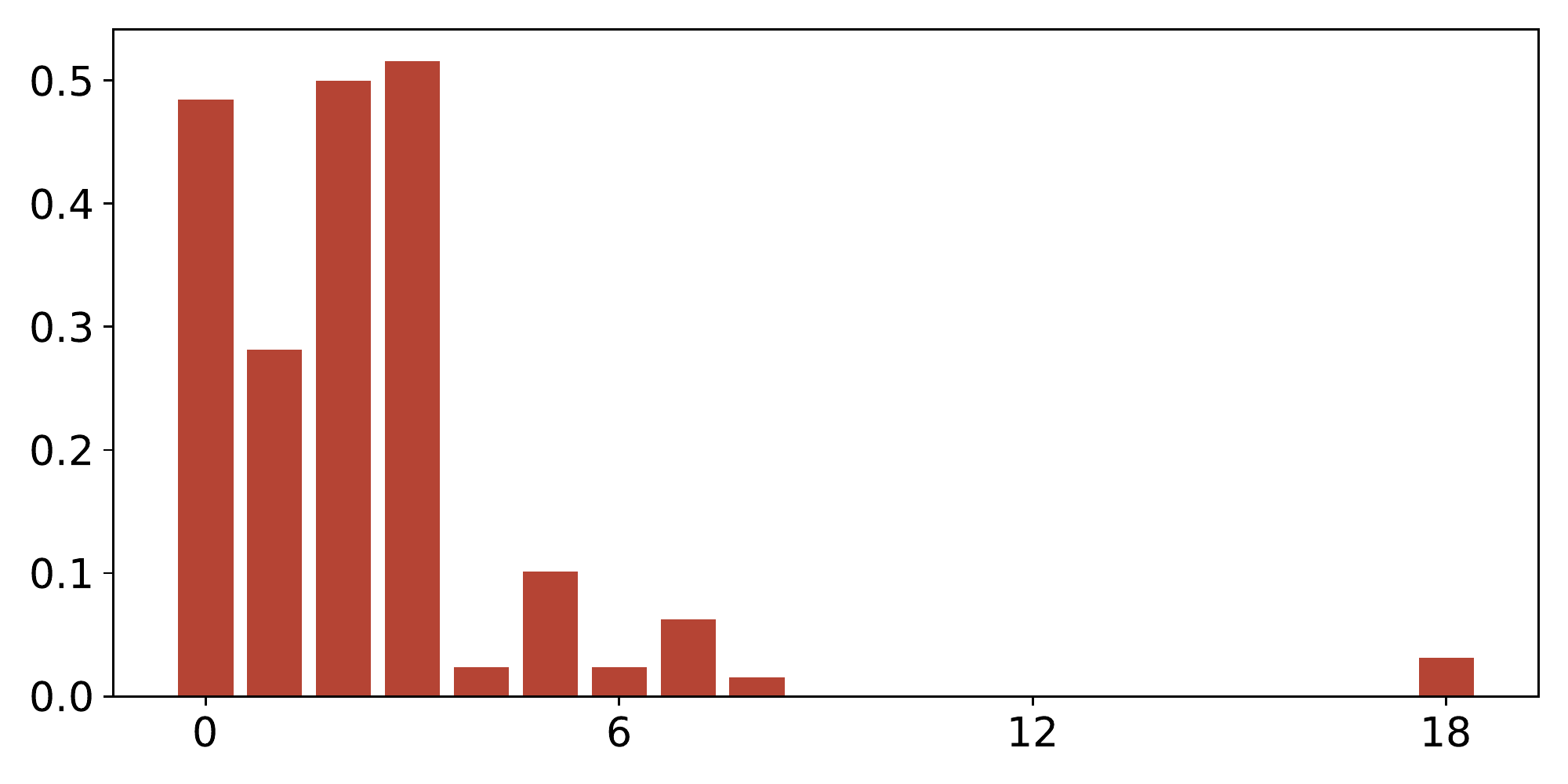}\\
\hspace{1.0em}
{\small{Index of Layers}}
\end{tabular}
\caption{Relative increment of size in each layer of a per-channel quantized ResNet-18 with multipoint quantization (W4A4).} \label{fig:neuron_increment_resnet18}
\vspace{-15pt}
\end{figure}

\begin{figure}[htbp]
\centering
\subfigure[Computation flowchart of typical dot product in a QNN]{
\begin{minipage}[b]{0.45\textwidth}
\includegraphics[width=.7\textwidth]{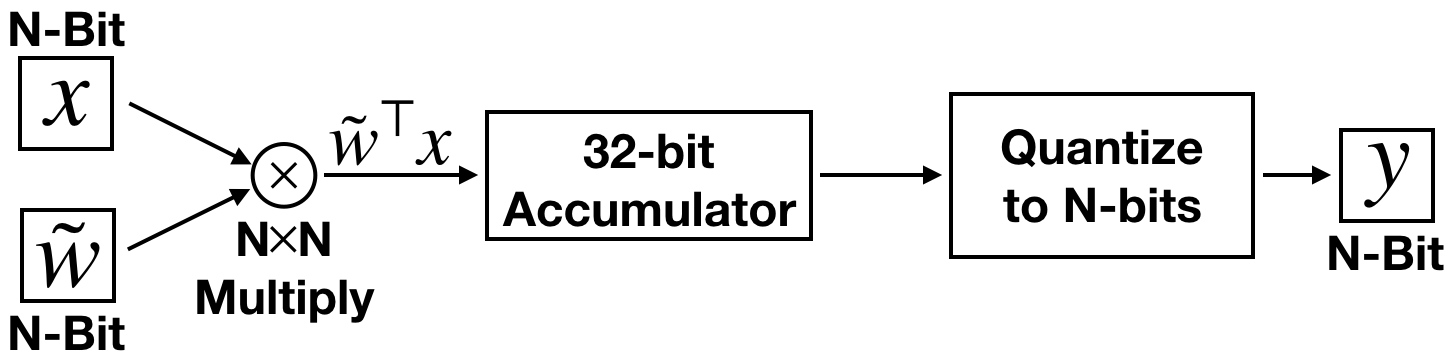}
\end{minipage}
}
\subfigure[Computation flowchart of dot product of multipoint quantization]{
\begin{minipage}[b]{0.48\textwidth}
\includegraphics[width=1\textwidth]{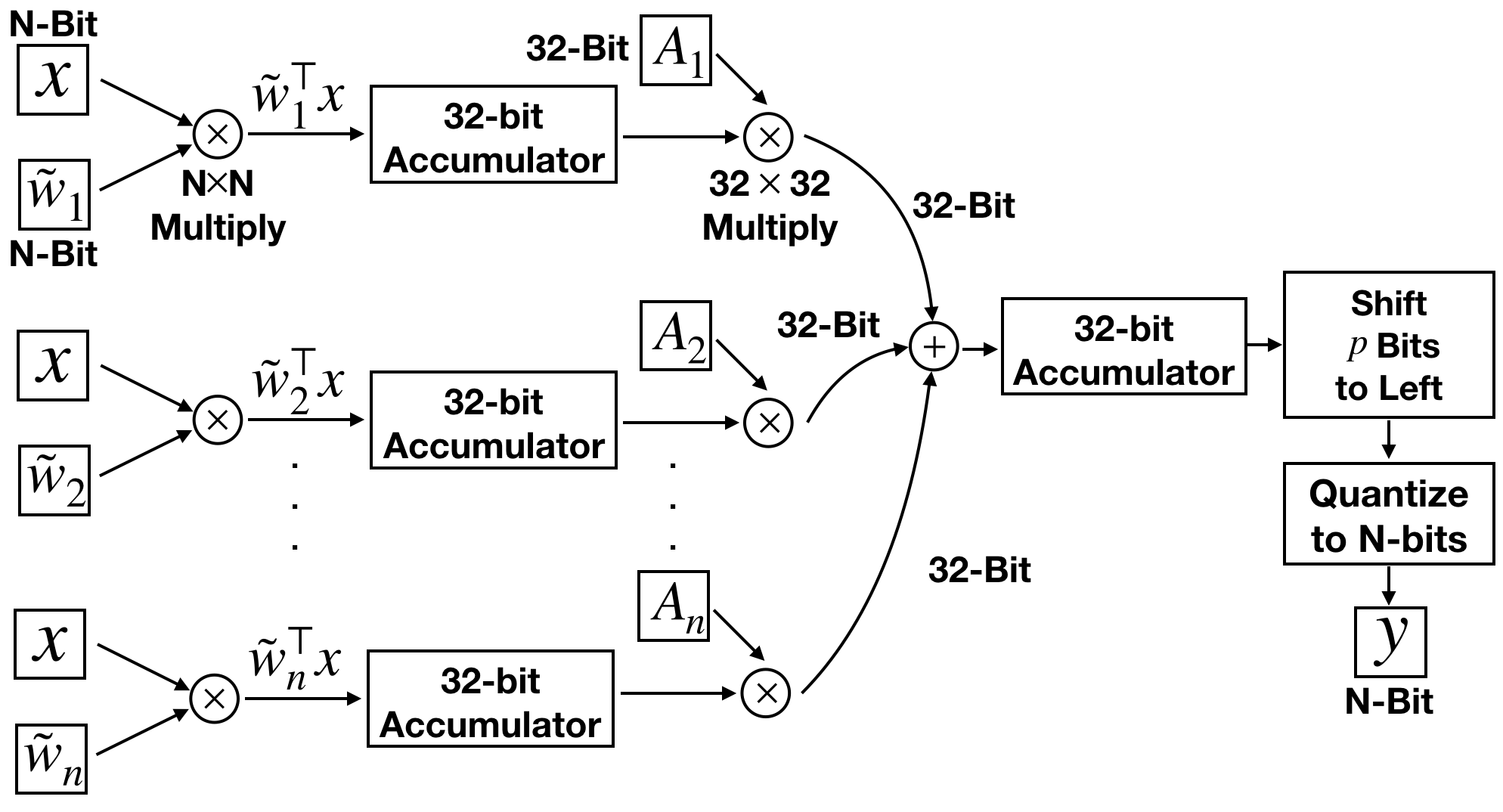}
\end{minipage}
}
\vspace{-10pt}
\caption{Flow charts of typical and multipoint quantization.} 
\vspace{-10pt}
\label{fig:hardware}
\end{figure}

\begin{figure}[htbp]
\centering
\renewcommand{\tabcolsep}{2pt}
\renewcommand\arraystretch{0.5}
\begin{tabular}{c}
\raisebox{3.0em}{\rotatebox{90}{\small{Error of Output}}}\includegraphics[width=0.46\textwidth]{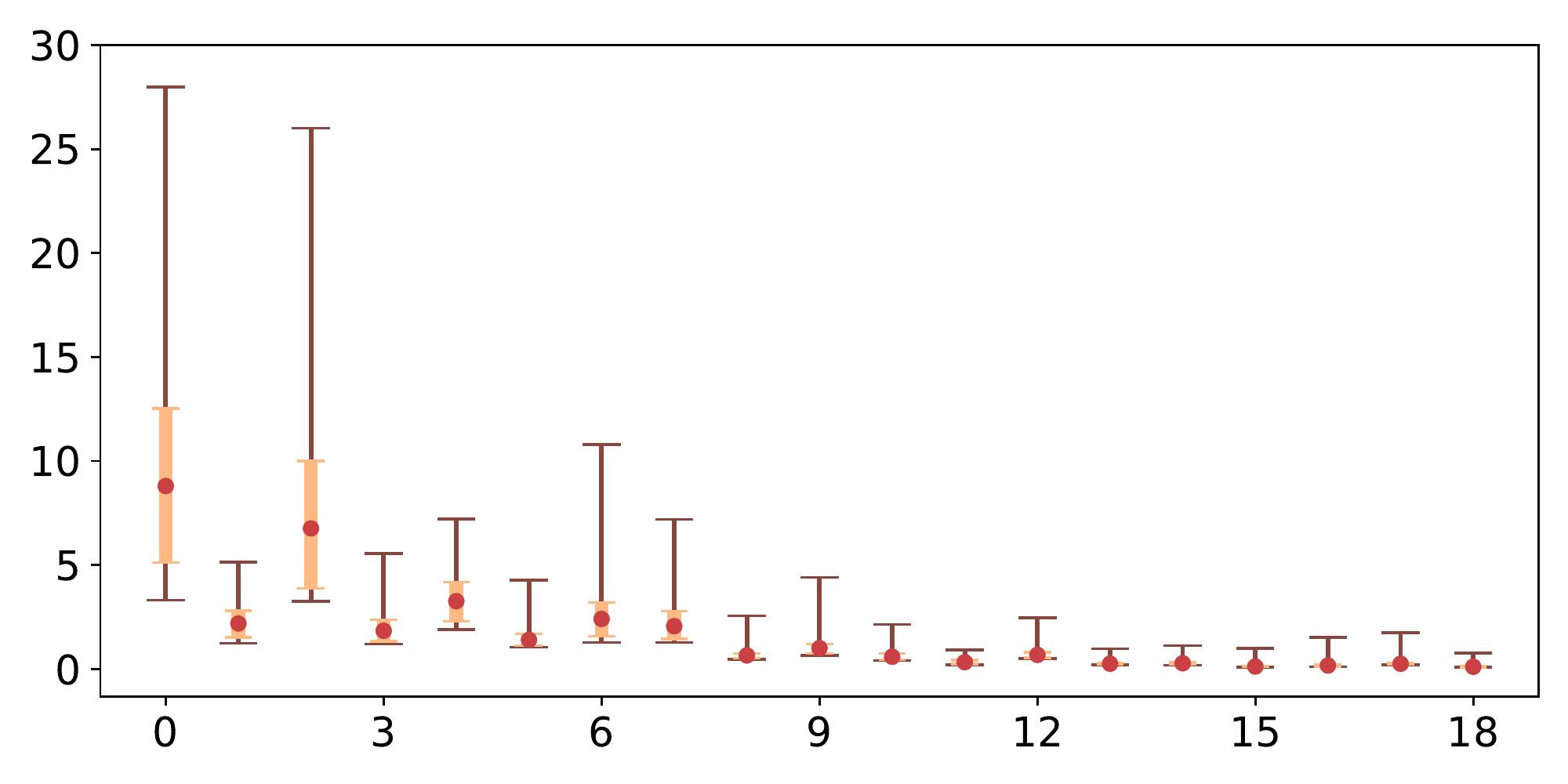}\\
\hspace{1.0em}
{\small{Index of Layers}}
\end{tabular}

\caption{The error plot of the output in a quantized ResNet-18. The red dot is the mean of output error of all channels in the corresponding layer. The dark bars show the maximum and minimum. The shallow region indicates the 15-th to 85-th percentile. Observations: (1) only a small portion of neurons have large error; (2) the starting layers are more sensitive to quantization.} \label{fig:error}
\vspace{-10pt}
\end{figure}
\end{document}